\documentclass{article}
\usepackage[utf8]{inputenc}
         
\usepackage{lmodern}
\usepackage{graphicx}      
\usepackage{natbib}
\usepackage{float}
\usepackage{color}
\usepackage{amsfonts}
\usepackage{amsmath}
\usepackage{amssymb}
\usepackage{geometry}
\usepackage{booktabs} 
\usepackage{array} 
\usepackage{paralist} 
\usepackage{verbatim} 
\usepackage{subfig}

\usepackage[titles,subfigure]{tocloft}

\DeclareMathOperator*{\E}{\mathbb{E}}

\begin{document}

 \begin{center}
 \Large
\textbf{Uncertainty quantification in a mechanical submodel driven by a Wasserstein-GAN \\}

\hspace{8pt}

\large
Hamza BOUKRAICHI$^{1,2}$, Nissrine AKKARI$^1$, \\ Fabien CASENAVE$^1$,  David RYCKELYNCK$^2$ \\

\hspace{8pt}

\small  
$^1$ Safran Tech \\
Etablissement Paris Saclay \\
Rue des Jeunes Bois-Chateaufort, 78114 Magny-Les-Hameaux, France\\ ~~ \\
$^2$ MINES ParisTech, PSL University \\
MAT - Centre des matériaux \\
CNRS UMR 7633, BP 87 91003 Evry, France
\end{center}

\par\rule{15cm}{0.4pt}
\begin{abstract} The analysis of parametric and non-parametric uncertainties of very large dynamical systems requires the construction of a stochastic model of said system. Linear approaches relying on random matrix theory \cite{Soize_2000} and principal componant analysis can be used when systems undergo low-frequency vibrations. In the case of  fast dynamics and wave propagation, we investigate a random generator of boundary conditions for fast submodels by using machine learning. We show that the use of non-linear techniques in machine learning and data-driven methods is highly relevant.

Physics-informed neural networks \cite{PINN}  is a possible choice for a data-driven method to replace linear modal analysis. An architecture that support a random component is necessary   for the construction of  the stochastic model of the physical system for  non-parametric uncertainties, since the goal is to learn the underlying probabilistic distribution of uncertainty in the data. Generative Adversarial Networks (GANs) are suited for such applications, where the Wasserstein-GAN with gradient penalty variant
 \cite{WGANP} offers improved convergence results for our problem.

The objective of our approach is to train a GAN on data from a finite element method code (Fenics) so as to extract stochastic
boundary conditions for faster finite element predictions on a submodel. The submodel and the training data have both the same geometrical support. It is a zone of interest for uncertainty quantification and relevant to engineering purposes. In the exploitation phase, the framework can be viewed as a randomized and parametrized
simulation generator on the submodel, which can be used as a Monte Carlo estimator. \\ \\

\end{abstract}
\par\rule{15cm}{0.4pt}

\section{Introduction}

The aim of this paper\footnote{\emph{This work has been submitted to IFAC for possible publication.}} is to present novel methods for submodeling using deep learning models for parametric and non-parametric uncertainty quantification for fast dynamics, where approaches based on linear modal analysis are computationally  inefficient and inaccurate.

\subsection{Related Work}
In order to   determine parametric and non-parametric approaches, one has to define the differences between aleatory and epistemic uncertainty. These laters are stated in \cite{3} 
and \cite{4} 
:
\begin{itemize}
\item Aleatory uncertainties: the uncertainties relative to some model parameters induced by the lack of
knowledge related to those parameters. To process these uncertainties, parametric approaches are used as the modeling of the uncertainty of the parameters by random variables and fields in order, for instance, to construct stiffness and mass matrices with respect to those parameters.
\item Epistemic uncertainties: also arise from lack of knowledge on parameters but  based on subjective perception, and limited data availability, such as interval analysis, possibility theory, and fuzzy set theory. Parametric approaches are not suited for this application.
\end{itemize}
Both these types of uncertainty are listed as model parameter uncertainty in \cite{3} where a new type of uncertainty is also introduced : 
\begin{itemize}
\item Modeling error: the uncertainties induced by the modeling errors within the choice of the physical model.
\end{itemize}
Epistemic uncertainties and modeling errors cannot be processed by fully parametric approches. Non-parametric and mixed approaches (see \cite{3}) are necessary such as: 
\begin{itemize}
\item Probabilistic approach: random matrix theory (see  \cite{1}, \cite{2})
\item Possibilistic approach: Fuzzy variables and interval analysis (see \cite{4}).   
\end{itemize}
In this paper, both parametric and non-parametric approaches are investigated in the situation of both aleatory and epistemic uncertainties. Modeling errors are not investigated. 

The use of neural networks to learn solutions of partial differential equations  (PDEs) have been recently proposed for physics application (see \cite{PINN} , \cite{PINN2}), using the real or an approximate of the  residual from the PDE to enforce a physical constraint on the output of the network. Such application exists for architectures like generative adversarial networks that were introduced in \cite{good2014} and optimized in \cite{WGANP}. More details on these architectures can be found in Section \ref{sct:Mod}. \\
 Generative adversarial networks and adversarial training are used for non-parametric density estimation in general cases of random data (\cite{abba_2019} and \cite{singh2018}) and also for physical data that are solutions of  certain partial differential equations (\cite{GAN-2019}).

Our study consists in using similar approaches to learn non-parametric densities over data from a finite element model, without any information about the underlying partial differential equation solved. But, we enforce physical properties using a submodel in the area of interest in the exploitation phase for uncertainty quantification.

\subsection{Contribution}
The aim of this paper is to present two novel methods developed  for the construction of stochastic submodels for uncertainty quantification using data from a finite element model (FEM). These two methods rely on the same general principle which is a stochastic submodel formed of two components:
\begin{itemize}
\item A neural network learning boundary conditions around a  predetermined zone of interest.
\item A finite element submodel in the zone of interest using boundary conditions generated by the neural network. We assume that there is no modeling error in the zone of interest covered by the proposed submodel.
\end{itemize}

The objective is to obtain comparable or/and better predictions than a classical learning process of a neural network over physical data, while improving some physical properties.
Indeed, during the training of a physics-informed neural network, increasing precision over physical properties is generally obtained using a penalization term given by the residual of the PDE in the cost function, but with FEM models designed for engineering applications, it is quite intrusive to get access to the residual of the PDE. \\

The development of deep neural networks that are thermodynamically-consistent is a key issue, as explained in \cite{HERNANDEZ2021}. In our approach, the known physical properties and principles are enforced, online,  using a submodel over the interest zone, here enforcing the output of the whole reduced model to be a solution to the underlying FEM formulation on the submodel zoom area. Also, the training of the neural networks is facilitated since every network learning problem is one dimension lower. For a 3D problem on a cartesian mesh, the network has to learn the prediction over a 2D surface representing the boundary conditions instead of learning the data over the whole 3D domain. \\

So to address both aleatory and epistemic uncertainties, we propose two methods as follows: 
\begin{itemize} 
\item Aleatory uncertainties: a deep convolutional neural regressor is trained to generate parametrized boundary conditions associated with the parameters of the simulation, for a parametric approach.
\item Epistemic uncertainties: a Wasserstein GAN is trained to generate stochastic boundary conditions by using  the same training data. It aims at learning the underlying probabilistic density binding the simulation data and the parameters of the simulation, for a non-parametric approach.
\end{itemize} 

Both methods are then compared to a linear data reduction, using the Proper Orthognal Decomposition (POD) method, constructed over the same boundary data.

\section{Models} \label{sct:Mod}
\subsection{Proper Orthogonal Decomposition (POD)}
Let us denote by $X=[L^{2}(\Omega)]$ the functional Hilbert space of the squared integrable scalar functions over a bounded $2D-$open set $\Omega$. We denote the $L^{2}(\Omega)$-inner product by $(.,.)$.

Consider $U(p)(t,x)\in \mathbb{R}$ the value of a physical data over a mesh of $\Omega$ and associated to the parameters vector $p$ and to a time $t$. The mesh has a grid shape, so that $U$ is also a tensor of data.  A subspace of the solution space is obtained thanks to the snapshots POD method~\cite{sirovich}: if we discretize the time interval to $m$ points, then the snapshots set is given as follows: $\mathcal{S}=\{U(p)(t_i)\;\;; i=1,...,m\}$.
The POD modes $\Phi_{j},$ $j=1,...,m$, computed via the snapshots POD start with the solution of the eigenvalues problem with the temporal correlations matrix:
\begin{equation}
 C_{ij}=(U(p)\left(t_i\right) , U(p)\left(t_j\right) ) ,
\end{equation}
of size $m \times m$. Let us denote by $(A_j)_{j=1,...,m}=\left(A_{i,j}\right)_{1\leq i\leq m}$ and $\left(\lambda_{j}\right)_{j=1,...,m}$ for
$j=1,...,M$, sets of respectively orthonormal eigenvectors and eigenvalues of the matrix~$C$. Then, the POD modes associated with
$\lambda_n$, are given by:
\begin{equation}\label{pod_velocity}
\Phi_{j}(x)=\frac{1}{\sqrt{\lambda_{n}}}\sum^{m}_{i=1}A_{i,j}U(p)(t_i,x) \;\;, \forall x \in\Omega \; \; \forall j=1,..., m.
\end{equation}
Snapshots are approximated by orthogonal projection on the space generated by a truncation of the POD basis: $U(p)(t,x)\approx \Sigma_{k=1}^{\hat{m}}\alpha_k(p,t)\Phi_{k}(x)$, where $\hat{m}\leq m$, and $\alpha_k$ are called the generalized coordinates. Meta-models are then trained to predict the generalized coordinates of a new solution from the parameter values.

\subsection{Deep Convolutional Neural Regressor}
A  Deep convolutional Neural Regressor (\textit{DcNR}) consists in learning to generate the physical data (U) over a grid with the parameters vector (p) as an input.  As indicated by its name, the internal structure of this network is formed by a succession of transposed convolutional layers of adequate dimensions in order to obtain a regression model of the physical field in the adequate size. The objective function in this case being:
\begin{equation}
\min\limits_\theta \E\limits_{p \in \mathbb{P}_{Train}} \sqrt{ [(U(p) - N(\theta,p)]^2 } 
\end{equation}
Where N denotes the neural network, $\theta$ its trainable weights, $ \mathbb{P}_{Train}$ the training set of parameters vectors, and $ \E $ the mathematical expectation.
\subsection{Wasserstein Generative Adversarial Network}
 Generative adversarial networks were introduced in \cite{good2014} as an unsupervised framework to learn probabilistic densities over data. It showed an empirical success as an efficient method for learning and sampling from a complicated multi-modal distribution. It relies on the adversarial training of two neural networks: 
\begin{itemize}
\item{Discriminator}: a classifier whose role is to determine whether the data it receives as inputs are real or generated by the second network.  Its architecture is a succession of  convolutional layers to determine a classifier network.
\item{Generator}: a generative model, whose role is to generate new data resembling the real data from a random vector (the input) in order to fool the discriminator. Its architecture is a succession of transposed convolutional layers for a generative model.
\end{itemize}
In \cite{WGANP}, it has been shown that under specific architecture and smoothness properties of the discriminator, an objective function defined as follows: 

\begin{equation}
 \min\limits_{\theta_{gen}} \max\limits_{\theta_{disc}} \E\limits_{z \sim \mathcal{N}(0,1)} [ D(\theta_{disc},G(\theta_{gen},z)) ] - \E\limits_{p \in  \mathbb{P}_{Train}}[ D(\theta_{disc},U(p)) ]
\label{eq:wganLoss}
\end{equation}
will lead the generator to convergence and being able to sample from the real data distribution using the random vector as a latent space descriptor.
In~\eqref{eq:wganLoss}, G and D denote respectively the generator and discriminator networks, and $\theta_{gen}$, $\theta_{disc}$ their respective trainable weights.
In the exploitation phase, the discriminator is no longer used and  the generator can be viewed as a randomized simulation generator on the submodel, which can be used as a Monte Carlo estimator.
\section{Use Case}\label{sct:UC}
\subsection{Domain Definition}
We define two 2D cartesian space grids $\Omega$ and $\Omega'$, with $ \Omega' \subset \Omega$ representing the zone of interest.
$\Omega$ and $\Omega '$ are space discretization of sizes $[N_x,N_y]$ and $[N'_x,N'_y]$ of the domains $[-L_x,L_x] \times  [-L_y,L_y]$  and  $[-L'_x,L'_x] \times  [-L'_y,L'_y]$. 
And finally a temporal grid $T$ is defined as discretization of size $N_T$ of the space $[0,T_{final}]$ and the time step $ \Delta t = \frac{T_{final}}{N_T-1}$ .

\subsection{Finite element models}
\begin{figure}[H]
\begin{center}
\includegraphics[scale=0.4]{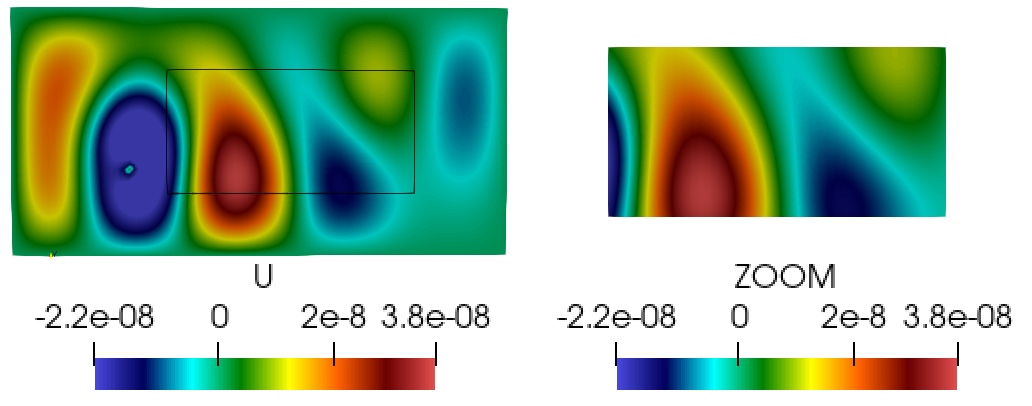}   
\caption{Visualization of the FEM output on $\Omega$ and  $ \Omega ^{'}$} 
\label{fig:FEM}
\end{center}
\end{figure}

The objective here is to train a generator on data from a FEM code (Fenics, see \cite{alnaes2015fenics}) so as to extract boundary values  for a submodel that occupies the zone of interest $ \Omega'$.
For visual representation of this approach, refer to Figure \ref{fig:FEM}.
Let g be the boundary values, g is defined as Dirichlet boundary conditions for both models as:
\begin{itemize}
\item For the initial FEM model, constant boundary values are chosen for the domain $\Omega$.
\item For the FEM submodel, g is the output of the pretrained neural network (the generator).
\end{itemize}

We choose to solve the 2D  wave equation, given as follows: 
\begin{equation} \left\{
\begin{array}{l}
  \frac{1}{c^2} \frac{\partial^2 u}{\partial t^2} - {\Delta u} = f ~~ on ~~  \Omega ~~ \forall t > 0  \\
u = g ~~   on  ~~ \partial  \Omega ~~ \forall t > 0  \\
 u = u_{0} ~~   on  ~~  \Omega ~~ for  ~~ t  =  0
\end{array}
\right.
\end{equation}
Where u is the amplitude of the wave (displacement on the z-axis).
For simplification purposes, the term $c^2$ will be omitted on the following formulation: \\ \\
The variational problem goes as: 
\begin{equation}\label{formul}  a(u^n,v) = L^n(v) \end{equation} 
Where $V_h$ is the Sobolev space of solutions on the approximate space. \\ \\
The time discretization used for the FEM formulation reads:\begin{equation} \frac{\partial^2 u}{\partial t^2}  =  \frac{u^n - 2 u^{n-1} + u^{n-2}}{\Delta t^2}\end{equation}
\begin{equation}\frac{u^n - 2 u^{n-1} + u^{n-2}}{\Delta t^2} = f^n + \Delta u^n \end{equation}
Then: 
\begin{equation} \forall v \in V_h  (u^n - 2 u^{n-1} + u^{n-2})v =\Delta t^2(f^n + \Delta u^n)v \end{equation}

\begin{equation}
 \int_{\Omega}u^n v \partial x - \int_{\Omega} \Delta t^2 \Delta u^n v \partial x =  \int_{\Omega} \Delta t^2 f^n + 2 u^{n-1}v - u^{n-2}v \partial x
\end{equation}

Using Green theorem:

\begin{equation} 
\int_{\Omega}u^n v \partial x +\Delta t^2  \int_{\Omega}  \Delta u^n \Delta v \partial x = \int_{\Omega} \Delta t^2 f^n + 2 u^{n-1}v - u^{n-2}v \partial x
\end{equation}

Then we obtain for the FEM formulation  (\ref{formul}): \\
 \begin{equation} a(u^n,v) =  \int_{\Omega}u^n v \partial x +\Delta t^2  \int_{\Omega}  \Delta u^n \Delta v \partial x\end{equation}
\begin{equation} L^n(v) =\int_{\Omega} \Delta t^2 f^n + 2 u^{n-1}v - u^{n-2}v \partial x \end{equation}

\subsection{Dataset Generation}\label{sct:data_gen}
A source point is determined for the problem resolution where $(x_S,y_S)$ are the source point coordinates, it is choosen to be outside the zoom domain:
 i.e.  $(x_S,y_S) \in \Omega$ and $(x_S,y_S) \not \in  \Omega'$. \\
The right hand side of the wave equation is set as: 
\begin{equation}   (\forall t \in T) 
 \left\{  
\begin{array}{l}
f(x_{S},y_{S},t) = sin(\omega t)  \\
  f(x,y,t) = 0 ~~ \forall (x,y) \in \Omega , (x,y) \ne (x_S,y_S)
\end{array}
\right.
\end{equation}

A three-dimensional parameter vector $  p = (\omega, x_s, y_s)$  is choosen and determined then sampled, (note that $c$ is fixed for all samples since it is a parameter needed for the submodel). 
Sampling is done using latin hypercube sampling routines. For every parameter vector $p$ a simulation matrix $U(p)$ is generated using FEM model described in section \ref{sct:UC}.
One sample of data is then$ (p,U(p))$ where $ p \in D_p \subset  \mathbb{R}^3 $ and $ U(p) \in V_h \subset \mathbb{R}^{N_X \times N_Y}.$  \\ 
Then, 3 datasets are generated as the following:
\begin{itemize}
\item Training data set: 100 samples generated, used for training each neural network described in section \ref{sct:Mod}.
\item Test data set:  10 samples generated, used for testing the training process of each neural network, and comparaison intra-model.
\item Monte Carlo samples: 1000 samples generated, used for uncertainty quantification and comparison of the estimate of the real probability density with the density from the neural networks.

\end{itemize}

\section{Numerical Results}\label{sct:NR}
\subsection{Data Sampling}
In this section we present the data range used for sampling and generating data for the training and testing phase.
Values were choosen as: $ L_x = 8m$ ,$ L_y = 4m$, $ L'_x =4m$, $L'_y=2m$ , $N_x = 40$, $N_y=20$, $N_T = 100$, $ \Delta t = 4 \times 10^{-5} s$, $c =2000 m/s $. Boundary conditions for the model over $ \Omega$ are set to be zero Dirichlet boundary conditions.

The variable parameters identified in Section \ref{sct:UC} are sampled  following Table \ref{tb:P_sample} values.

\begin{table}[H]
\begin{center}
\caption{Parameters sampling}\label{tb:P_sample}
\begin{tabular}{ccccc}
P & Mean Value & Variation (\%) & Min Value & Max Value \\ \hline
$\omega$ & 5 kHz & 5\% & 4,75 kHz & 5,25 kHz \\
$x_S$  & -1.85 m &    17.5\% of $ L_y$  & -2.2 m  & -1.5 m  \\ 
 $y_S$  & -0.65 m &   28,75\% of $ L_y$   & -1.8 m  & 0.5  m \\ 
\hline
\end{tabular}
\end{center}
\end{table}
\subsection{Trained submodels}
For every model described in Section \ref{sct:Mod}, we train multiple version in order to do a full comparison for the two approaches:
\begin{itemize}
\item \textbf{POD}: We train different POD models with multiple metamodels over the orthogonal projection coefficients (random forest, gaussian process, linear ...). We choose to keep a POD model with random forest considering it held the best trade-off between precision and computional cost for our problem. It will be referred to as \textit{POD\_RF}.
\item \textbf{DcNR}: We  train multiple DcNR : 
	\begin{itemize}
	  \item \textit{NN}: it takes as input the parameter vector p and outputs the value of U over all the area of interest.
	  \item \textit{NN\_BC}:  it takes as input the parameter vector p and outputs the boundary values around  the area of interest.
	  \item \textit{NN\_t}:  it takes as input the parameter vector p and the time value t  and outputs the value of U over all the area of interest at the instant t.
	  \item \textit{NN\_BC\_t}:   it takes as input the parameter vector p and the time value t  and outputs the boundary values around  the area of interest at the instant t.
	\end{itemize}  
\item \textbf{GAN}:  Like for the DcNR, we trained two versions, both taking as an input a random vector z and outputs the value of U over all the area of interest (\textit{WGAN}) or the boundary values around  the area of interest  (\textit{WGAN\_BC}) . Predictions of \textit{WGAN} and \textit{WGAN\_BC} restricted to the boundary of $\Omega'$ are also applied as Dirichlet boundary conditions to the submodel (\textit{WGAN\_ZOOM} and \textit{WGAN\_BC\_ZOOM} respectively).
\end{itemize}
For information about training time of each neural network, refer to Table \ref{tb:tt}.
\begin{table}[H]
\begin{center}
\caption{Training time}\label{tb:tt}
\begin{tabular}{ccc}
Nets & Training time & GPU card \\ \hline
NN & 12,4 Hours & NVIDIA V100  \\
NN\_BC  & 2,7 Hours &  NVIDIA V100  \\
WGAN & 24 Hours & NVIDIA A100  \\
WGAN\_BC  & 7,8 Hours &  NVIDIA A100  \\    
\hline
\end{tabular}
\end{center}
\end{table}

We define a relative error indicator over the time and space grid of the interest zone, in order to quantify the precision of our submodels as $\epsilon$.
For a submodel M, a parameter vector p (resp. random vector z for a GAN), and a time value t:
\begin{equation}
\epsilon (M,p,t) =\frac{\E\limits_{(x,y) \in \Omega '} [ \displaystyle\left\lvert M(p)(t,x,y)  - U(p)(t,x,y) \right\rvert ]}{\max\limits_{x,y \in \Omega '} \displaystyle\left\lvert U(p)(t,x,y) \right\rvert } 
\end{equation}

For a comparison over the testing data set: 
\begin{equation}
\epsilon (M,t) =\E\limits_{p \in \mathbb{P}_{Test}} [\epsilon (M,p,t)] ~ or ~ \E\limits_{z \sim \mathcal{N}(0,1)} [\epsilon (M,z,t)] 
\end{equation}

For a comparison of the prediction of physical quantities we choose to compute the kinetic energy over the zone of interest grid using a finite difference scheme as follows:
\begin{equation}
K_e(p,t,x,y) =\frac{ m}{2} \left( V(p,t,x,y) \right) ^2  
\end{equation}
Where:
\begin{equation}
V(p,t,x,y) = \frac{ U(p)(t,x,y) -  U(p)(t-dt,x,y)}{dt} 
\end{equation}

Since the mass (m) is constant over the space grid and over all parameter vectors, it will be omitted in computing the relative error  over the kinetic energy prediction.

\subsection{Parametric approach results}
For the parametric approach, comparison is done by computing the error indicator defined in the previous section for all our parametric submodels, for all the samples in the testing data set described in Section \ref{sct:UC}.
\begin{figure}[H]
\begin{center}
\includegraphics[scale=0.3]{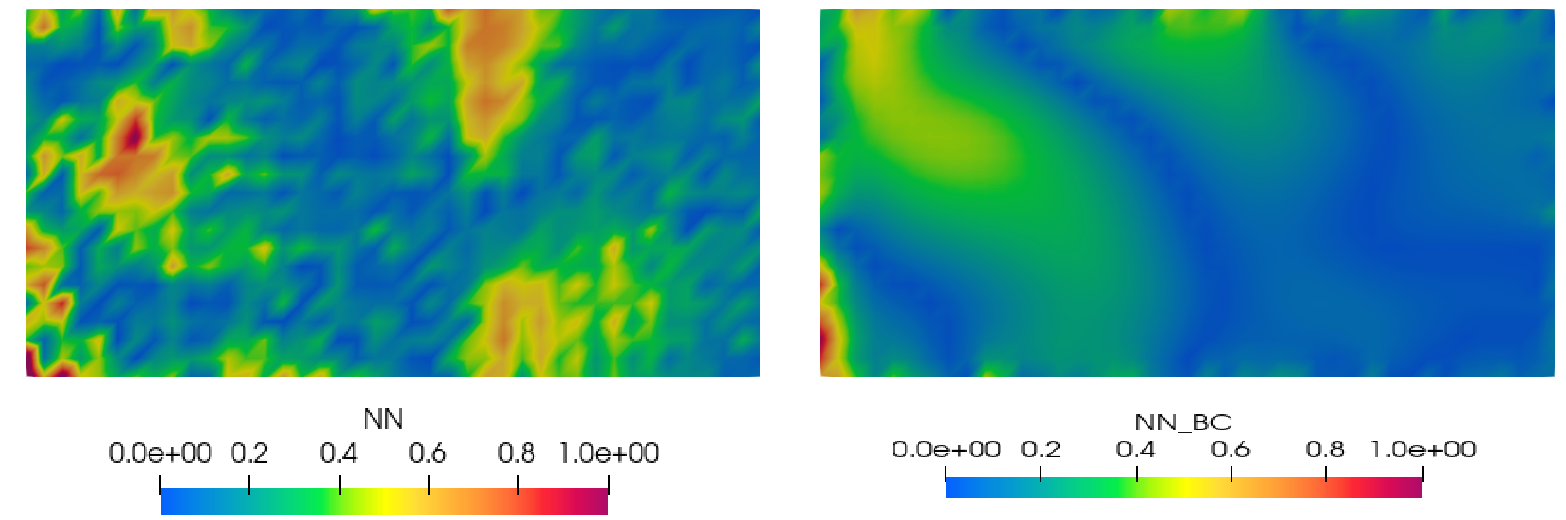}   
\caption{Relative error $\epsilon$:  NN (Left) vs NN-Zoom (Right)  } 
\label{fig:err_pix}
\end{center}
\end{figure}

\begin{figure}[H]
\begin{center}
\includegraphics[scale=0.5]{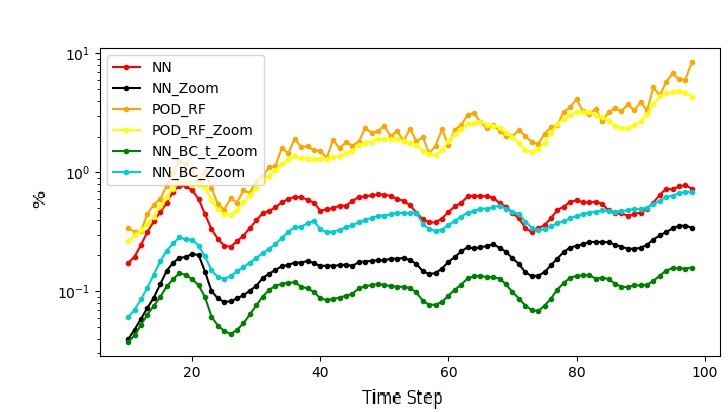}   
\caption{Relative error $\epsilon$ on $K_e$} 
\label{fig:f3}
\end{center}
\end{figure}
 Figure \ref{fig:err_pix} shows one of the known problems with using convolutional layers to predict physical fields, which is errors and noise introduced in the output following the structure of the different convolutions, this phenomenom is corrected by the zoom operation by the submodel. As shown, the noise is still visible on the boundaries but not propagated inside the interest area.
 Figure \ref{fig:f3} shows that the submodels approaches are better in predicting physical values such as kinetic energy, this can be explained by the fact that the submodels consists in running a partial physical model and thus having better physical properties, and as expected the POD performs poorly against non-linear methods.  
\subsection{Non-parametric approach results}
For the non-parametric approach, since comparison on regression models is impossible, we used our submodels as Monte-Carlo estimators of statistical quantities and compared the estimated values with the same Monte-Carlo approach on the real data. We choose to estimate the mean and compute the error indicator defined in the previous section. And to evaluate the generative capacity of our models, we define a discrepancy indicator as follows:
\begin{equation}
\sigma(M,t,x,y) = \sqrt{\E\limits_{z \sim \mathcal{N}(0,1) } [( M(z)(t,x,y) - \mathbb{E}_{Train})^{2}]}
\end{equation}
Where M is a WGAN-based network and $ \mathbb{E}_{Train}$ is the pointwise mean over the training data: 
\begin{equation}
 \mathbb{E}_{Train} = \E\limits_{p \in \mathcal{P}_{Train}} [U(p)(t,x,y)]
\end{equation}
$\sigma$ computes a point wise discrepancy value to show our models capacity to generate different data from the training data. To evaluate this generative capacity we define a relative discrepancy indicator as follows:
\begin{equation}
\sigma_{rel}(M,t) = \frac{  \E\limits_{(x,y) \in \Omega '}  [ \displaystyle\left\lvert \sigma(M,t,x,y)  -  \sigma_{Train} \right\rvert ]}{\max\limits_{x,y \in \Omega '} \sigma_{Train}}
\end{equation}
Where $\sigma_{Train}$ is the pointwise standard deviation over the training data: 
\begin{equation}
 \sigma_{Train} =\sqrt{ \E\limits_{p \in \mathcal{P}_{Train}} [(U(p)(t,x,y) - \mathbb{E}_{Train})^{2}]}
\end{equation}
We choose as physical value the maximum amplitude defined as follows:
\begin{equation}
A(p)(x,y) = \displaystyle\left\lvert \max\limits_{t} U(p)(t,x,y) -  \min\limits_{t} U(p)(t,x,y) \right\rvert
\end{equation}
\begin{figure}[H]
\begin{center}
\includegraphics[scale=0.3]{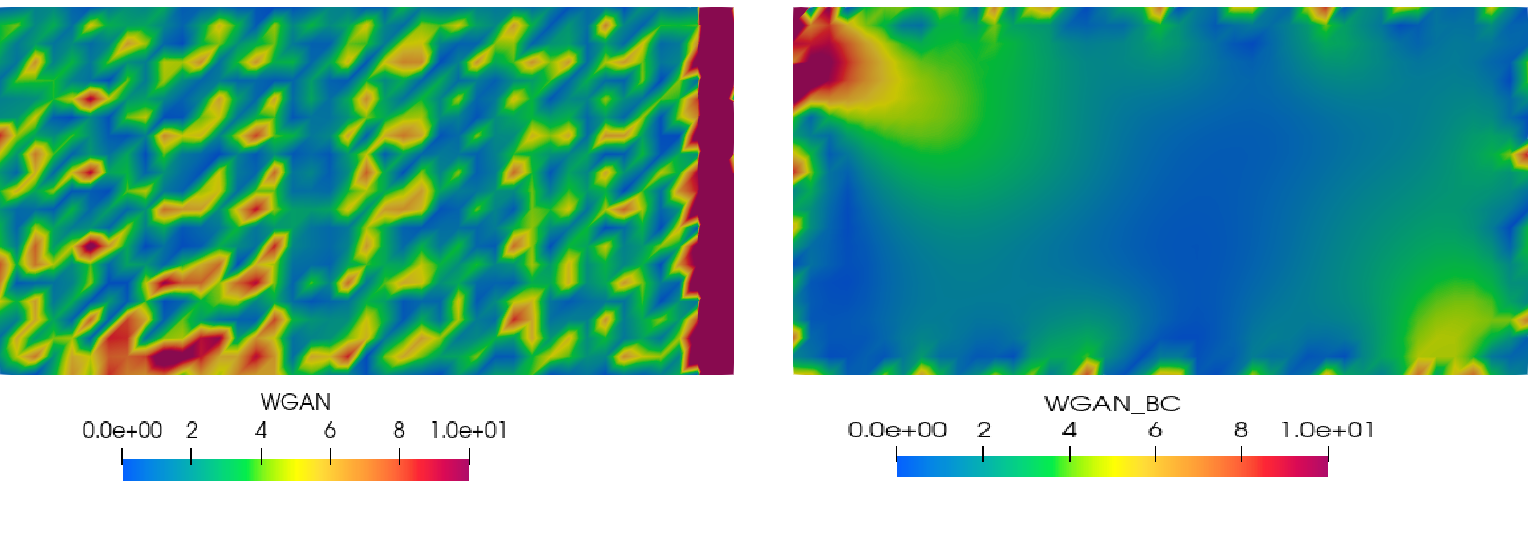}   
\caption{Relative error $\epsilon$ on pointwise mean:  GAN (Left) and GAN-BC-ZOOM (Right)  } 
\label{fig:mean_gan}
\end{center}
\end{figure}

\begin{figure}[H]
\begin{center}
\includegraphics[width=12.5cm,height=9cm]{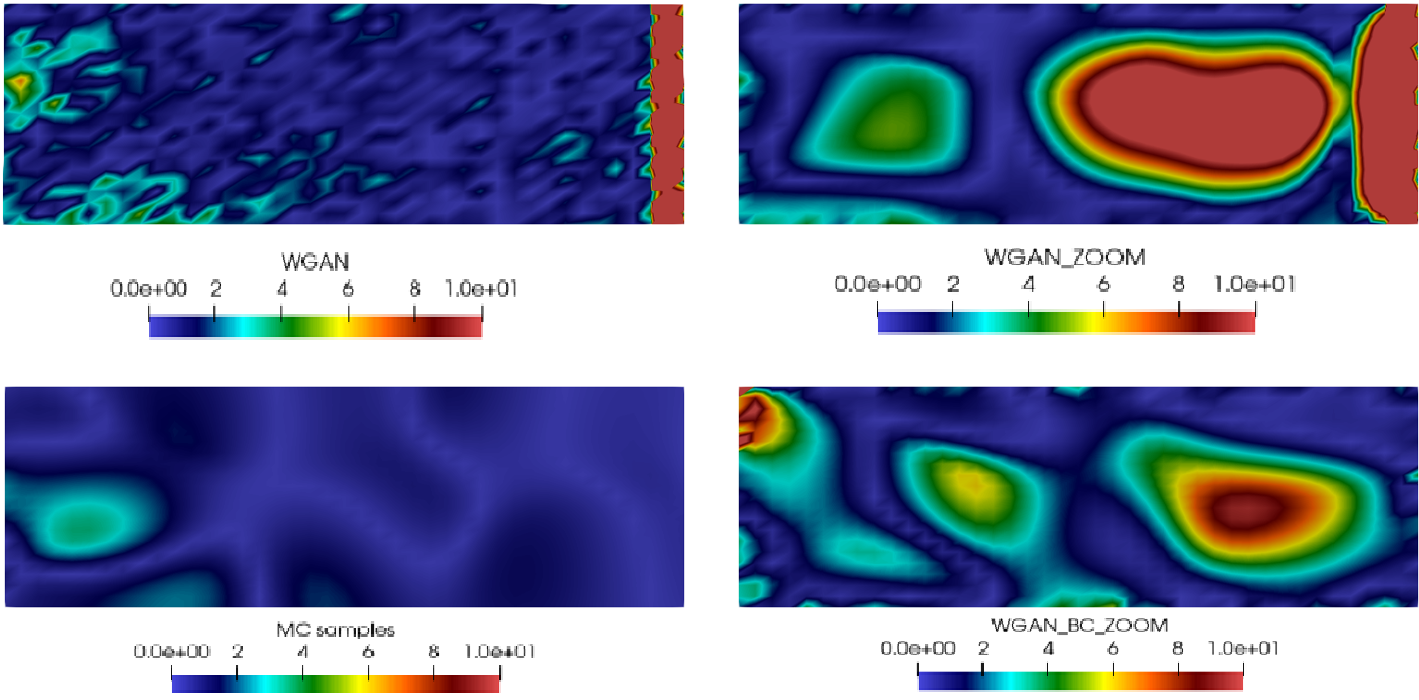}   
\caption{Point wise relative discrepancy indicator $\sigma_{rel}$ :  WGAN (Upper left) , WGAN-ZOOM (Upper right), Monte Carlo samples (Lower left) and GAN-BC-ZOOM (Lower right)  } 
\label{fig:std_gan}
\end{center}
\end{figure}

\begin{figure}[H]
\begin{center}
\includegraphics[scale=0.5]{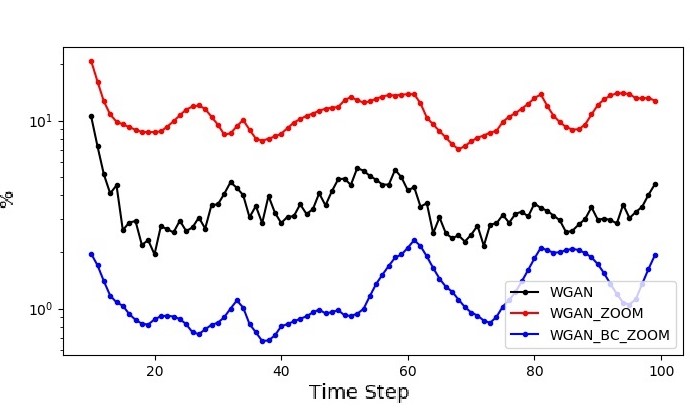}   
\caption{Relative error $\epsilon$ of WGANs prediction on pointwise mean} 
\label{fig:gans_mean}
\end{center}
\end{figure}

\begin{figure}[H]
\begin{center}
\includegraphics[scale=0.5]{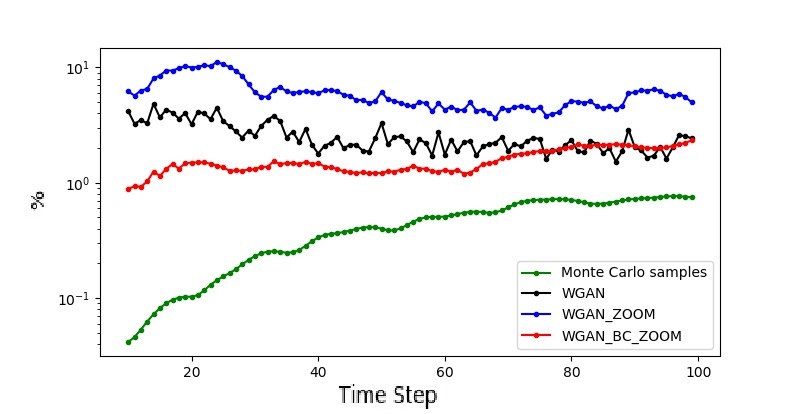}   
\caption{Point wise relative discrepancy indicator $\sigma_{rel}$ of WGANs} 
\label{fig:std_error}
\end{center}
\end{figure}

\begin{figure}[H]
\begin{center}
\includegraphics[scale=0.5]{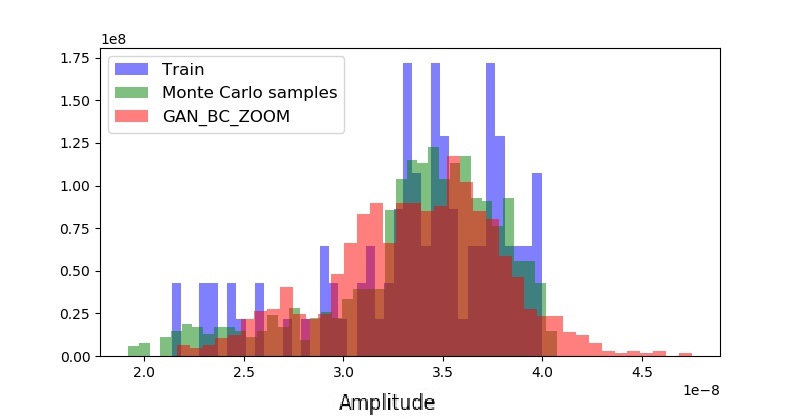}   
\caption{Histogram of maximum amplitude prediction} 
\label{fig:amp}
\end{center}
\end{figure}

Figure \ref{fig:mean_gan}   shows that the zoom operation of the submodel helds the same correction properties over the noise and errors introduced by the use of convolutional layers and also noise introduced by the GAN's random component. Figure \ref{fig:gans_mean} shows that our submodel approaches perfom significantly better on the mean prediction.
Figures~\ref{fig:std_gan}  and~\ref{fig:std_error} show that our approaches held better generative properties that are useful for uncertainty quantification, the $WGAN\_BC\_ZOOM$ shows the best
 performance, since higher discrepancy values in the other approaches can be explained by the accumulation of the error on the right boundary side of the submodel zoom area, furthermore the discrepancy in the $WGAN\_BC\_ZOOM$ approach shows a more structured shape holding more statistically representative physical information. Figures \ref{fig:std_gan}  and  \ref{fig:std_error} show also the discrepancy indicator computed on the 1000 Monte Carlo samples described in Section~\ref{sct:UC}, the low value can be explained by the fact that the size of samples in training set is sufficent to describe the underlying probabilistic distribution of the Monte Carlo samples knowing the parameters. Further investigations are necessary to precisely determine the sample size of the training set for better generative behavior. Nonetheless, our approaches show better generative capacity exploring extremum values that have not been considered in the training set and the Monte Carlo samples as shown in Figure~\ref{fig:amp} where our approach has better generative capacity on the density tails.

\section{Conclusion}\label{sct:cl}

In this paper we presented novel methods for parametric and non-parametric uncertainty quantification relying on physical submodels over an area of interest. We have empirically shown that our methods obtain comparable and slighlty better estimation of physical fields than classical neural networks approaches, while reducing the dimensionality of the learning problem and thus reducing the training cost of our models by restricting our attention to the boundary of a submodel. We fulfill  the necessary condition that the cost of each run of the physical submodel is smaller than the cost of running the full physical model. Better precision is reached in the parametric view, by using DCnR's. Besides, in situation where the parameters distribution is unknwon (epistemic uncertainties), only non-parametric approaches are feasible. For that, using the Wasserstein-GAN as a boundary conditions generator, we showed a higher value of the discrepancy in the Monte Carlo sampling method compared to high-fidelity solutions, while keeping physical consistency thanks to the learned boundary conditions, thus offering better generative behavior in the exploration of density tails.

\bibliographystyle{agsm}
\bibliography{ref}        

@article{1, title={A reduced-order random matrix approach for stochastic structural dynamics}, volume={88}, number={21-22}, journal={Computers \& Structures}, author={Adhikari, S. and Chowdhury, R.}, year={2010}, pages={1230-1238}}

@article{2, title={Robustness of structural reliability analyses to epistemic uncertainties}, volume={28}, journal={Mechanical Systems and Signal Processing}, author={Guedri, M. and Cogan, S. and Bouhaddi, N.}, year={2012}, pages={458-469}}

@article{3, title={Model identification in computational stochastic dynamics using experimental modal data}, volume={50-51}, journal={Mechanical Systems and Signal Processing}, author={Batou, A. and Soize, C. and Audebert, S.}, year={2015}, pages={307-322}}

@article{4, title={A new structural reliability analysis method in presence of mixed uncertainty variables}, volume={33}, number={6}, journal={Chinese Journal of Aeronautics}, author={You, Lingfei and Zhang, Jianguo and Du, Xiaosong and Wu, Jie}, year={2020}, pages={1673-1682}}

@article{soize_2000, title={A nonparametric model of random uncertainties for reduced matrix models in structural dynamics}, volume={15}, number={3}, journal={Probabilistic Engineering Mechanics}, author={Soize, C.}, year={2000}, pages={277-294}}

@article{PINN,
      title={Physics Informed Deep Learning (Part I): Data-driven Solutions of Nonlinear Partial Differential Equations}, 
      author={Maziar Raissi and Paris Perdikaris and George Em Karniadakis},
      year={2017},
      eprint={1711.10561},
      archivePrefix={arXiv},
      primaryClass={cs.AI}
}

@misc{WGANP,
      title={Improved Training of Wasserstein GANs}, 
      author={Ishaan Gulrajani and Faruk Ahmed and Martin Arjovsky and Vincent Dumoulin and Aaron Courville},
      year={2017},
      eprint={1704.00028},
      archivePrefix={arXiv},
      primaryClass={cs.LG}
}

@article{PINN2, title={Physics-informed neural networks: A deep learning framework for solving forward and inverse problems involving nonlinear partial differential equations}, volume={378}, journal={Journal of Computational Physics}, author={Raissi, M. and Perdikaris, P. and Karniadakis, G.e.}, year={2019}, pages={686-707}}

@article{GAN-2019, title={Adversarial uncertainty quantification in physics-informed neural networks}, volume={394}, journal={Journal of Computational Physics}, author={Yang, Yibo and Perdikaris, Paris}, year={2019}, pages={136-152}}

@article{good2014,
  title={Generative adversarial nets},
  author={Goodfellow, Ian and Pouget-Abadie, Jean and Mirza, Mehdi and Xu, Bing and Warde-Farley, David and Ozair, Sherjil and Courville, Aaron and Bengio, Yoshua},
  journal={Advances in neural information processing systems},
  volume={27},
  year={2014}
}

@article{abba_2019, title={A Generative Adversarial Density Estimator}, journal={2019 IEEE/CVF Conference on Computer Vision and Pattern Recognition (CVPR)}, author={Abbasnejad, M. Ehsan and Shi, Qinfeng and Hengel, Anton Van Den and Liu, Lingqiao}, year={2019}}

@article{singh2018,
  title={Nonparametric density estimation under adversarial losses},
  author={Singh, Shashank and Uppal, Ananya and Li, Boyue and Li, Chun-Liang and Zaheer, Manzil and P{\'o}czos, Barnab{\'a}s},
  journal={arXiv preprint arXiv:1805.08836},
  year={2018}
}

@article{alnaes2015fenics,
  title={The FEniCS project version 1.5},
  author={Aln{\ae}s, Martin and Blechta, Jan and Hake, Johan and Johansson, August and Kehlet, Benjamin and Logg, Anders and Richardson, Chris and Ring, Johannes and Rognes, Marie E and Wells, Garth N},
  journal={Archive of Numerical Software},
  volume={3},
  number={100},
  year={2015}
}

@article{HERNANDEZ2021,
title = {Deep learning of thermodynamics-aware reduced-order models from data},
journal = {Computer Methods in Applied Mechanics and Engineering},
volume = {379},
pages = {113-763},
year = {2021},
issn = {0045-7825},
author = {Quercus Hernandez and Alberto Badías and David González and Francisco Chinesta and Elías Cueto}
}

@article{sirovich,
	Author = {L. Sirovich},
	Journal = {Part III: dynamics and scaling. Quarterly of applied mathematics},
	pages = {583-590},
	Title = {Turbulence and the dynamics of coherent structures},
	Volume = {45},
	Year = {1987}}
                                                                   
\end{document}